\documentclass[pmlr]{jmlr}

\usepackage{longtable}

\usepackage{booktabs}
\usepackage[load-configurations=version-1]{siunitx} 

\theorembodyfont{\upshape}
\theoremheaderfont{\scshape}
\theorempostheader{:}
\theoremsep{\newline}

\jmlrvolume{126}
\jmlryear{2020}
\jmlrworkshop{Machine Learning for Healthcare}

\makeatletter
\def\set@curr@file#1{\def\@curr@file{#1}} 
\makeatother
\usepackage[load-configurations=version-1]{siunitx}

\usepackage{microtype}
\usepackage{graphicx}
\usepackage{subcaption}
\usepackage{booktabs}
\usepackage{amsmath}
\usepackage{color,soul}
\usepackage{algorithm}
\usepackage{algorithmic}
\usepackage{multirow}

\def \bg {\mathbf{b}}
\def \ins {\mathbf{i}}

\title[Deep RL for Blood Glucose Control]{Deep Reinforcement Learning for\\  Closed-Loop Blood Glucose Control}

\author{\Name{Ian Fox} 
        \Email{ifox@umich.edu}\\ 
        \addr Department of Computer Science Engineering\\
       University of Michigan\\
       Ann Arbor, MI, USA
       \AND
       \Name{Joyce Lee}\\
       \addr Department of Pediatrics and Communicable Diseases\\
       University of Michigan\\
       Ann Arbor, MI, USA
       \AND
       \Name{Rodica Pop-Busui}\\
       \addr Department of Internal Medicine\\
       University of Michigan\\
       Ann Arbor, MI, USA
       \AND
       \Name{Jenna Wiens}
       \Email{wiensj@umich.edu}\\
       \addr Department of Computer Science Engineering\\
       University of Michigan\\
       Ann Arbor, MI, USA} 

\begin{document}

\maketitle

\begin{abstract}
  People with type 1 diabetes (T1D) lack the ability to produce the insulin their bodies need. As a result, they must continually make decisions about how much insulin to self-administer to adequately control their blood glucose levels. Longitudinal data streams captured from wearables, like continuous glucose monitors, can help these individuals manage their health, but currently the majority of the decision burden remains on the user. To relieve this burden, researchers are working on closed-loop solutions that combine a continuous glucose monitor and an insulin pump with a control algorithm in an `artificial pancreas.' Such systems aim to estimate and deliver the appropriate amount of insulin. Here, we develop reinforcement learning (RL) techniques for automated blood glucose control. Through a series of experiments, we compare the performance of different deep RL approaches to non-RL approaches. We highlight the flexibility of RL approaches, demonstrating how they can adapt to new individuals with little additional data. On over 2.1 million hours of data from 30 simulated patients, our RL approach outperforms baseline control algorithms: leading to a  decrease in median glycemic risk of nearly 50\% from 8.34 to 4.24 and a decrease in total time hypoglycemic of 99.8\%, from  4,610 days to 6. Moreover, these approaches are able to adapt to predictable meal times (decreasing average risk by an additional 24\% as meals increase in predictability). This work demonstrates the potential of deep RL to help people with T1D manage their blood glucose levels without requiring expert knowledge. All of our code is publicly available, allowing for replication and extension.\footnote{https://gitlab.eecs.umich.edu/mld3/rl4bg}
\end{abstract}
\section{Introduction}
Type 1 diabetes (T1D) is a chronic disease affecting 20-40 million people worldwide \citep{you_type_2016}, and its incidence is increasing \citep{tuomilehto_emerging_2013}. People with T1D cannot produce insulin, a hormone that controls blood glucose levels, and must monitor their blood glucose levels and manually administer insulin doses as needed. Administering too little insulin can lead to hyperglycemia (high blood glucose levels), which results in chronic health complications \citep{control_resource_1995}, while administering too much insulin results in hypoglycemia (low blood glucose levels), leading to coma or death. Getting the correct dose requires careful measurement of glucose levels and carbohydrate intake, resulting in at least 15-17 data points a day. When using a continuous glucose monitor (CGM), this can increase to over 300 data points, or a blood glucose reading every 5 minutes \citep{coffen_magnitude_2009}.

CGMs with an insulin pump, a device that delivers insulin, can be used with a closed-loop controller as an `artificial pancreas' (AP). Though the technology behind CGMs and insulin pumps has advanced, there remains significant room for improvement when it comes to the control algorithms \citep{bothe_use_2013, pinsker_randomized_2016}. Current hybrid closed-loop approaches require accurate meal announcements to maintain glucose control.

In this work, we investigate deep reinforcement learning (RL) for blood glucose control. RL is a promising solution, as it is well-suited to learning complex behavior and readily adapts to changing domains \citep{clavera2018learning}. We hypothesize that deep RL, the combination of RL with a deep neural network, will be able to accurately infer latent meals to control insulin. Furthermore, as RL is a learning-based approach, we hypothesize that RL will adapt to predictable meal schedules better than baseline approaches. 

The fact that RL is learning-based means it requires data to work effectively. Unlike many other health settings, there are credible simulators for blood glucose management \citep{visentin_university_2014}. Having a simulator alleviates many concerns of applying RL to health problems \citep{gottesman2018evaluating, gottesman_guidelines_2019}. However, that does not mean RL for blood glucose control is straightforward, and, in this paper, we identify and address several challenges. To the best of our knowledge, we present the first deep RL approach that achieves human-level performance in controlling blood glucose without requiring meal announcements. 

\subsection*{Generalizable Insights about Machine Learning in the Context of Healthcare}
Applying deep RL to blood glucose management, we encountered challenges broadly relevant for RL in healthcare. As such, we believe our solutions and insights, outlined below, are broadly relevant as well.
\begin{itemize}
    \item The range of insulin and carbohydrate requirements across patients makes it difficult to find a single action space that balances the needs of rapid insulin administration and safety. Indeed, many health problems involve changes in action distributions across patients (\textit{e.g.} in anesthesia dosing \citep{bouillon_size}). To solve this problem, we present a robust patient-specific action space that naturally encourages safer policies. 
    \item We found several pitfalls in evaluating our proposed approach that led to unrealistic performance estimates. To address this issue, we used validation data to perform careful model selection, and used extensive test data to evaluate the quality of our models. In RL, it is typical to report performance on the final trained model (without model selection) over a handful of rollouts. Our experiments demonstrate the danger of this approach.
    \item Deep RL has been shown to be unstable \citep{rlblogpost, henderson_deep_2018}, often achieving poor worst-case performance. This is unacceptable for safety-critical tasks, such as those in healthcare. We found that a combination of simple and widely applicable approaches stabilized performance. In particular, we used a safety-augmented reward function, realistic randomness in training data, and random restarts to train models that behaved safely over thousands of days of evaluation.
    \item Finally, unlike game settings where one has ability to learn from hundreds of thousands of hours of interaction, any patient-specific model must be able to achieve strong performance using a limited amount of data. We show that a simple transfer learning approach can be remarkably sample efficient and can even surpass the performance of models trained from scratch. 
\end{itemize}

\section{Background and Related Work}
This work develops and applies techniques in reinforcement learning to the problem of blood glucose management. To frame this work, we first provide a brief introduction to RL, both in general and specifically applied to problems in healthcare. We then discuss how RL and other approaches have been used for blood glucose control and present an overview on blood glucose simulation.

\subsection{Reinforcement Learning}\label{sec:rl}
RL is an approach to optimize sequential decision making in an environment, which is typically assumed to follow a Markov Decision Process (MDP). An MDP is characterized by a 5-tuple $(S, A, P, R, \gamma)$, where $s \in S$ are the states of the environment, $a \in A$ are actions that can be taken in the environment, the transition function $P: (s, a) \rightarrow s'$ defines the dynamics of the environment, the reward function $R: (s, a) \rightarrow r \in \mathbb{R}$ defines the desirability of state-action pairs, and the discount factor $\gamma \in [0, 1]$ determines the tradeoff between the value of immediate and delayed rewards. The goal in RL is to learn a policy $\pi: s \rightarrow a$, or function mapping states to actions, that maximizes the expected cumulative reward, or:
\begin{equation}
    \arg\max_{\pi \in \Pi} \sum_{t=1}^{\infty} E_{s_t \sim P(s_{t-1}, a_{t-1})}[\gamma^t R(s_t, \pi(s_t))],
\end{equation}
where $\Pi$ is the space of possible policies and $s_0 \in S$ is the starting state. The state value function, $V(s)$, is the expected cumulative reward where $s_0 = s$. The state-action value function $Q(s,a) = R(s,a) + E_{s' \sim P(s, a)}[\gamma V(s')]$ extends the notion of value to state-action pairs.

\subsection{Reinforcement Learning in Healthcare}
In recent years, researchers have started to explore RL in healthcare. Examples include matching patients to treatment in the management of sepsis \citep{weng_representation_2017, komorowski_artificial_2018} and mechanical ventilation \citep{prasad_reinforcement_2017}. In addition, RL has been explored to provide contextual suggestions for behavioral modifications \citep{klasnja_efficacy_2019}. Despite its successes, RL has yet to be fully explored as a solution for a closed-loop AP system \citep{bothe_use_2013}. 

\subsection{Algorithms for Blood Glucose Control}
Among recent commercial AP products, proportional-integral-derivative (PID) control is the most common backbone \citep{trevitt_artificial_2015}. The simplicity of PID controllers make them easy to use, and in practice they achieve strong results \citep{steil2013algorithms}. The Medtronic Hybrid Closed-Loop system, one of the few commercially available, is built on a PID controller \citep{garg_glucose_2017, ruiz_effect_2012}. A hybrid closed-loop controller adjusts baseline insulin rates but requires human intervention to control for the effect of meals. The main weakness of PID controllers is their reactivity. As a result, they often cannot react fast enough to meals, and thus rely on meal announcements \citep{garg_glucose_2017}. Additionally, without safety modifications, PID controllers can deliver too much insulin, triggering hypoglycemia \citep{ruiz_effect_2012}. In contrast, we hypothesize that an RL approach will be able to leverage patterns associated with meal times, resulting in more responsive and safer policies.

Previous works have examined RL for different aspects of blood glucose control. See \cite{tejedor_reinforcement_2020} for a recent survey. Many of these works investigated the use of RL to adapt existing insulin treatment regimens to learn a ‘human-in-the-loop’ policy \citep{ngo_reinforcement-learning_2018, oroojeni_mohammad_javad_reinforcement_2015, sun_dual_2018}. This contrasts with our setting, where we aim to learn a fully closed-loop policy.

Like our work, \cite{daskalaki_preliminary_2010} and \cite{de_paula_-line_2015} focus on the task of closed-loop glucose control. \cite{daskalaki_preliminary_2010} use direct future prediction to aid PID-style control, substituting the problem of RL with prediction. \cite{de_paula_-line_2015} use a policy-iteration framework with Gaussian process approximation and Bayesian active learning. While they tackle a similar problem, these works use a simple simulator and a fully deterministic meal routine for training and testing. In our experiments, we use an FDA-approved glucose simulator and a realistic non-deterministic meal schedule, significantly increasing the challenge.

\subsection{Glucose Models and Simulation}
Models of the glucoregulatory system are important for the development and testing of an AP \citep{cobelli_integrated_1982}. In our experiments, we use the UVA/Padova model \citep{kovatchev_silico_2009}. This simulator models the glucoregulatory system as a nonlinear multi-compartment system, where glucose is generated in the liver, absorbed through the gut, and controlled by external insulin. A more detailed explanation can be found in \cite{kovatchev_silico_2009}. For reproducibility, we use an open-source version of the simulator that comes with 30 virtual patients \citep{xie_simglucose_2018}. The parameter distribution of the patient population is determined by age, and the simulator comes with 10 children, adolescents, and adults each \cite{kovatchev_silico_2009}. We combine the simulator with a non-deterministic meal schedule (\textbf{Appendix \ref{app:meal}}) to realistically simulate patient behavior.

\section{Methods}
We present a deep RL approach well suited for blood glucose control. In framing our problem, we pay special attention to the concerns of partial observability and safety. The issue of partial observability motivates us to use a maximum entropy control algorithm, soft actor-critic, combined with a recurrent neural network. Safety concerns inspire many aspects of our experimental setup, including our choice of action space, reward function, and evaluation metrics. We also introduce several strong baselines, both with and without meal announcements, to which we compare.

\subsection{Problem Setup}\label{ssec:setup}
We frame the problem of closed-loop blood glucose control as a partially-observable Markov decision process (POMDP) consisting of the 7-tuple $(S^*, O, S, A, P, R, \gamma)$. A POMDP generalizes the MDP setting described in \textbf{Section \ref{sec:rl}} by assuming we do not have access to the true environment states, here denoted $s^* \in S^*$, but instead observe noisy states $s \in S$ according to the (potentially stochastic) observation function: $O: s^* \rightarrow s$. This setting applies given the noise inherent in CGM sensors \citep{vettoretti2019modeling} and our assumption of unobserved meals.

In our setting, the true states $\mathbf{s}^*_t \in S^*$ are the 13-dimensional simulator states, as described in \cite{kovatchev_silico_2009}. The stochastic observation function $O: \mathbf{s}^*_t \rightarrow b_t, i_t$ maps from the simulator state to the CGM observation $b_t$ and insulin $i_t$ administered. To provide temporal context, we augment our observed state space $\mathbf{s}_t \in S$ to include the previous 4 hours of CGM $\bg^t$ and insulin data $\ins^t$ at 5-minute resolution: $\mathbf{s}_t = [\bg^t, \ins^t]$ where:
$$
    \bg^t = [b_{t-47}, b_{t-46}, \dots, b_{t}], 
    \ins^t = [i_{t-47}, i_{t-46}, \dots, i_{t}],
$$
$b_{t} \in \mathbb{N}_{40:400}$, $i_{t} \in \mathbb{R}_{+}$, and $t \in \mathbb{N}_{1:288}$ represents a time index for a day at 5-minute resolution. Note that in our augmented formulation, the observation function no longer directly maps to observed states, as observed states incorporate significant amounts of historical data. We chose a 4-hour window, as we empirically found it led strong performance. We use a time resolution of 5 minutes to mimic the sampling frequency of many common CGMs. Actions $a_t \in \mathbb{R}_{\ge 0}$ are real positive numbers, denoting the size of the insulin bolus in medication units. 

The transition function $P$, our simulator, consists of two elements: i) $M: t \rightarrow c_t$ is the meal schedule, and is defined in \textbf{Appendix \ref{app:meal}}, and ii) $G: (a_t, c_t, \mathbf{s}^*_t) \rightarrow (b_{t+1}, i_{t+1}, \mathbf{s}^*_{t+1})$, where $c_t \in \mathbb{R}_{\ge 0}$ is the amount of carbohydrates input at time $t$ and $G$ is the UVA/Padova simulator \citep{kovatchev_silico_2009}. Note that our meal schedule is patient-specific, and includes randomness in the daily number, size, and timing of meals.

\begin{figure}
    \centering
    \includegraphics[width=0.45\linewidth]{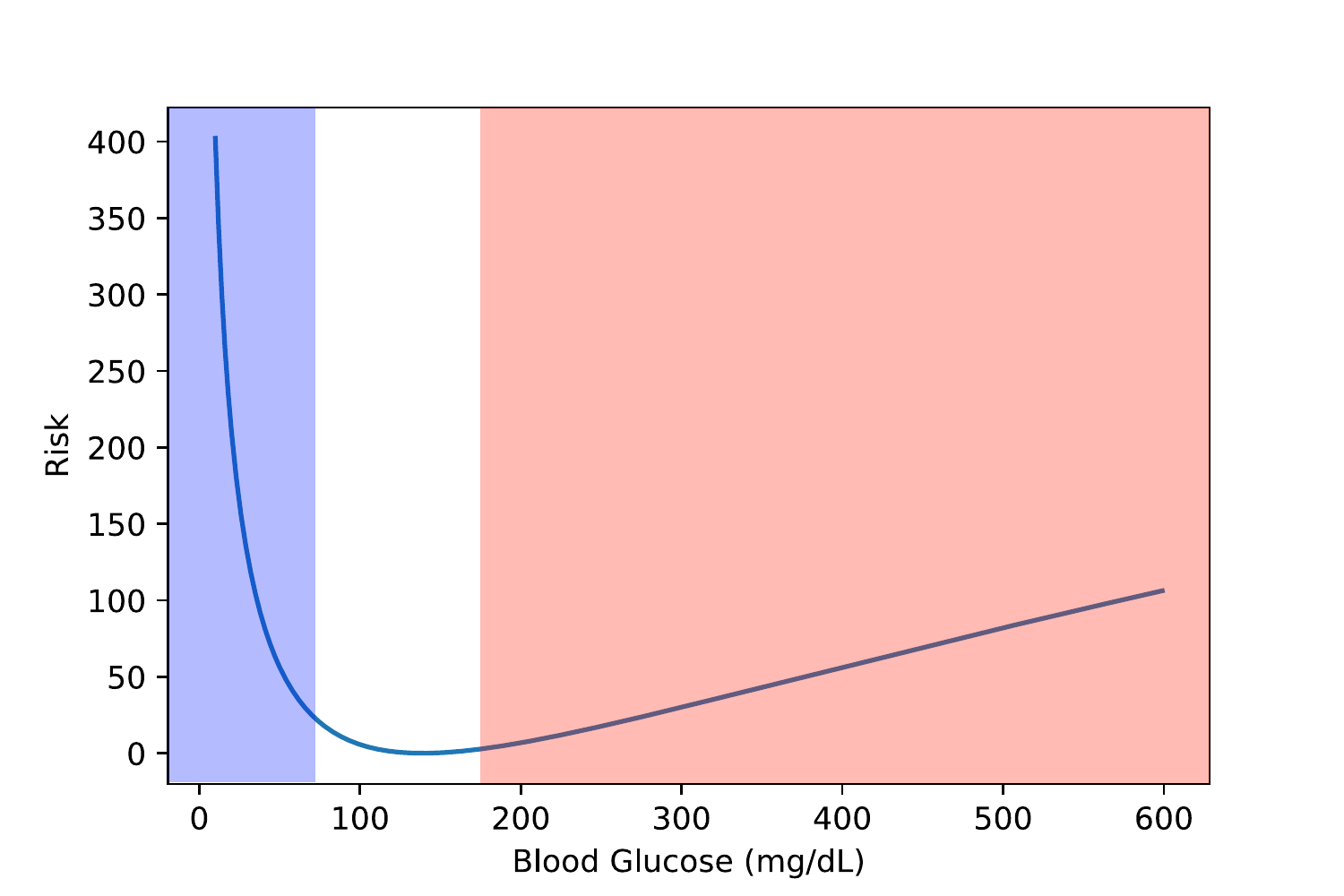}
    \caption{The risk function proposed in \citep{magni_model_2007}. The mapping between blood glucose values (in mg/dL, x-axis) and risk values (y-axis). Blood glucose levels corresponding to hypoglycemia are shown in the blue shaded region, the glucose range corresponding to hyperglycemia is shown in the red shaded region. This function identifies low blood glucose values as higher risk than high blood glucose values, which is sensible given the rapidity of hypoglycemia.}
    \label{fig:risk}
\end{figure}

The reward function $R: (\mathbf{s}_t, a_t) \rightarrow \mathbb{R}$ is defined as negative $-risk(b_t)$ where $risk$ is the Magni risk function:
\begin{equation}
    risk(b) = 10*(c_0 * \log(b)^{c_1}-c_2)^2,
\end{equation}
$c_0=3.35506$, $c_1=0.8353$, and $c_2=3.7932$ \citep{magni_model_2007}. These values are set such that $risk(70)=risk(280)=25$, see \textbf{Figure \ref{fig:risk}}. Finally, we set $\gamma=0.99$ for our experiments, a value we determined empirically on validation data in combination with the early termination penalty.

Considered in isolation, our reward could lead to dangerous behavior. As it is always negative, cumulative reward is maximized by ending the episode as quickly as possible, which occurs when glucose reaches unsafe levels. To avoid this, we add a termination penalty of $-1e5$ to trajectories that enter dangerous blood glucose regimes (blood glucose levels less than 10 or more than 1,000 mg/dL), countering the advantage of early termination. We investigated other reward functions, such as time in range or distance from a target blood glucose value, but found this reward worked best. It led to control schemes with less hypoglycemia, as low blood glucose is penalized more heavily than high glucose. Low glucose can occur quickly when large amounts of insulin are given without an accompanying meal. Given the lack of meal announcements and sensor noise in our setting, avoiding hypoglycemia was a significant challenge. 

\subsection{Soft Actor-Critic}
We chose to use the soft actor-critic (SAC) algorithm to learn glucose control policies. We initially experimented with a Deep-Q Network approach \citep{mnih_human-level_2015}. However, choosing a discretized action space (as is required by Q-learning) that accounted for the range of insulin values across a day and allowed exploration proved impractical, as large doses of inappropriately timed insulin can be dangerous. Among continuous control algorithms, we selected SAC as it has been shown to be sample efficient and competitive \citep{haarnoja_soft_2018}. Additionally, maximum entropy policies like the ones produced by SAC can do well in partially observed settings like our own \citep{eysenbach2019if}.

SAC produces a stochastic policy $\pi: \mathbf{s}_t \rightarrow p(a)$ $\forall a \in A$, which maps a state to a distribution over possible actions. Under SAC, the policy (or actor) is represented by a neural network with parameters $\phi$. Our network generates outputs $\mu, \log(\sigma)$ which parameterize a normal distribution $\mathcal{N}(\mu, \sigma)$. The actions are distributed according to a TanhNormal distribution, or $tanh(z), z \sim \mathcal{N}(\mu, \sigma)$. $\pi_\phi$ is trained to maximize the maximum entropy RL objective function:
\vspace{-1pt}
\begin{equation}\label{eqn:policy_return}
    J(\pi) = \sum_{t=0}^T \mathbb{E}_{(\mathbf{s}_t, a_t) \sim P(\mathbf{s}_{t-1}, \pi_\phi(\mathbf{s}_{t-1}))} [R(\mathbf{s}_t, a_t) + \alpha H(\pi_\phi(\cdot|\mathbf{s}_t))],
\end{equation}

\vspace{-1pt}
where entropy, $H$, is added to the expected cumulative reward to improve exploration and robustness \citep{haarnoja_soft_2018}. Intuitively, the return in \textbf{Equation \ref{eqn:policy_return}} encourages a policy that can obtain a high reward under a variety of potential actions. The temperature hyperparameter $\alpha$ controls the tradeoff between reward and entropy. In our work, we set this using automatic temperature tuning \citep{haarnoja_soft_2018a}. \textbf{Equation \ref{eqn:policy_return}} is optimized by minimizing the KL divergence between the action distribution and the distribution induced by the state-action values:
\vspace{-1pt}
\begin{equation}\label{eqn:pi}
    J_{\pi}(\phi)=\mathbb{E}_{\mathbf{s}_{t} \sim \mathcal{D}}\left[\mathrm{D}_{\mathrm{KL}}\left(\pi_{\phi}\left(\cdot | \mathbf{s}_{t}\right) \| \frac{\exp \left(Q_{\theta}\left(\mathbf{s}_{t}, \cdot\right)\right)}{Z_{\theta}\left(\mathbf{s}_{t}\right)}\right)\right]
\end{equation}

\vspace{-1pt}
where $\mathcal{D}$ is a replay buffer containing previously seen $(\mathbf{s}_t, \mathbf{a}_t, r_t, \mathbf{s}_{t+1})$ tuples, $Z_\theta$ is a partition function, and $Q_\theta$ is the state-action value function parameterized by a neural network (also called a critic). This means that our learned policy engages in probability matching, selecting an action with probability proportional to its expected value. This requires an accurate value function. To achieve this, $Q_\theta$ is trained by minimizing the temporal difference loss:
\begin{gather}\label{eqn:q}
    J_{Q}(\theta)=\mathbb{E}_{\left(\mathbf{s}_{t}, \mathbf{a}_{t}\right) \sim \mathcal{D}}\left[\frac{1}{2}\left(Q_{\theta}\left(\mathbf{s}_{t}, \mathbf{a}_{t}\right)-\hat{Q}\left(\mathbf{s}_{t}, \mathbf{a}_{t}\right)\right)^{2}\right], \\
    \hat{Q}\left(\mathbf{s}_{t}, \mathbf{a}_{t}\right)=r\left(\mathbf{s}_{t}, \mathbf{a}_{t}\right)+\gamma \mathbb{E}_{\mathbf{s}_{t+1} \sim p}\left[V_{\overline{\psi}}\left(\mathbf{s}_{t+1}\right)\right].
\end{gather}
\vspace{-1pt}
$V_\psi$ is the soft value function parameterized by a third neural network, and $V_{\overline{\psi}}$ is the running exponential average of the weights of $V_\psi$ over training. This is a continuous variant of the hard target network replication in \citep{mnih_human-level_2015}. $V_\psi$ is trained to minimize:
\begin{equation}\label{eqn:v}
    J_{V}(\psi)=\mathbb{E}_{\mathbf{s}_{t} \sim \mathcal{D}}\left[\frac{1}{2}\left(V_{\psi}\left(\mathbf{s}_{t}\right)-\mathbb{E}_{\mathbf{a}_{t} \sim \pi_{\rho}}\left[Q_{\theta}\left(\mathbf{s}_{t}, \mathbf{a}_{t}\right)-\log \pi_{\phi}\left(\mathbf{a}_{t} | \mathbf{s}_{t}\right)\right]\right)^{2}\right].
\end{equation}

In summary: we learn a policy that maps from states to a probability distribution over actions, the policy is parameterized by a neural network $\pi_\phi$. Optimizing this network (\textbf{Equation \ref{eqn:pi}}) requires an estimation of the soft state-action value function, we learn such an estimate $Q_\theta$ (\textbf{Equation \ref{eqn:q}}) together with a soft value function $V_\psi$ (\textbf{Equation \ref{eqn:v}}). Additional details of this approach, including the gradient calculations, are given in \citep{haarnoja_soft_2018}. In keeping with previous work, when testing our policy we remove the sampling component, instead selecting the mean action $tanh(\mu)$. We replace the MSE temporal difference loss in \textbf{Equation} \ref{eqn:q} with the Huber loss, as we found this improved convergence.

\subsubsection{Recurrent Architecture}\label{sec:architecture}
Our network $\pi_\phi$ takes as input the past 4 hours of CGM and insulin data, requiring no human input (\textit{i.e.}, no meal announcements). To approximate the true state $\mathbf{s}^*_t$ from the augmented state $s$ we parameterize $Q_\theta$, $V_\psi$, and $\pi_\phi$ using gated-recurrent unit (GRU) networks \citep{cho_learning_2014}, as GRUs have been successfully used for glucose modeling previously \citep{fox_deep_2018, zhu_deep_nodate}.

\subsubsection{Patient-Specific Action Space} 
After the network output layer, actions are squashed using a \textit{tanh} function. Note that this results in half the action space corresponding to negative values, which we interpret as administering no insulin. This encourages sparse insulin utilization and makes it easier for the network to learn to safely control baseline glucose levels. To ensure that the maximum amount of insulin delivered over a 5-minute interval is roughly equal to a normal meal bolus for each individual, we use the average ratio of basal to bolus insulin in a day \citep{kuroda_basal_2011} to calculate a scale parameter for the action space, $\omega_{b}=43.2*bas$, where $bas$ is the default patient-specific basal insulin rate provided by \cite{xie_simglucose_2018}.

\subsection{Efficient Policy Transfer} 
 One of the main disadvantages of deep RL is sample efficiency. Thus, we explored transfer learning techniques to efficiently transfer existing models to new patients. We refer to our method trained from scratch as RL-Scratch, and the transfer approach as RL-Trans. For RL-Trans, we initialize $Q_\theta, V_\psi$ and $\pi_\phi$ for each class of patients (children, adolescents, and adults) using fully trained networks from one patient of that source population (see \textbf{Appendix \ref{app:patient}}). We then fine-tune these networks on data collected from the target patient. 

 When fine-tuning, we modify the reward function by removing the termination penalty and adding a constant positive value (100) to all rewards. This avoids the negative reward issue discussed in \textbf{Section \ref{ssec:setup}}. Removing the termination penalty increased the consistency of returns over training, allowing for a more consistent policy gradient. The additional safety provided by a termination penalty is not required as we begin with policies that are already stable. We found this simple approach for training patient-specific policies attains good performance while using far less patient-specific data.
 
\subsection{Baselines}
We examine three baseline methods for control: basal-bolus (\textbf{BB}), \textbf{PID} control, and PID with meal announcements (\textbf{PID-MA}). BB reflects an idealized human-in-the-loop control strategy, and PID reflects a common closed-loop AP algorithm. PID with meal announcements is based on current AP technology, and is a combination of the two, requiring regular human intervention. Finally, we consider an 'oracle' approach that has access to the true state $s^*_t$. This decouples the task of learning a policy from state inference, serving as a pseudo-upper bound on performance for our proposed approach.

\subsubsection{Basal-Bolus Baseline}
This baseline is designed to mimic human control and is an ideal depiction of how an individual with T1D controls their blood glucose. In this setting, we modify the state representation $\mathbf{s}_t$ to include a carbohydrate signal and a cooldown signal (explained below), and to remove the historical data, $\mathbf{s}_t = [b_t, i_t, c_t, cooldown]$. Note that the inclusion of a carbohydrate signal, or meal announcement, places the burden of providing accurate and timely estimates of meals on the person with diabetes. Each virtual patient in the simulator comes with the parameters necessary to calculate a reasonable basal insulin rate $bas$ (the same value used in our action space definition), correction factor $CF$, and carbohydrate ratio $CR$. These three parameters, together with a glucose target $b_g$, define a clinician-recommended policy $\pi(s_t) = bas + (c_t > 0) * (\frac{c_t}{CR} + cooldown * \frac{b_t - b_g}{CF})$ where $cooldown$ is 1 if there have been no meals in the past three hours, otherwise it is 0. The cooldown ensures that each meal is only corrected for once. Appropriate settings for these parameters can be estimated by endocrinologists, using previous glucose and insulin information \citep{walsh_guidelines_2011}. The parameters for our virtual patient population, which are derived from a distribution validated by clinical trials \citep{kovatchev_silico_2009}, are given in \textbf{Appendix \ref{app:bb}}.

\subsubsection{PID Baseline}
PID controllers are a common and robust closed-loop baseline \citep{steil2013algorithms}. A PID controller operates by setting the control variable, $a_t$, to the weighted combination of three terms $a_t = k_P P(b_t) + k_I I(b_t) + k_D D(b_t)$ such that the process variable $b_t$ ($t$ is the time index) remains close to a specified setpoint $b_g$. The terms are calculated as follows: i) the proportional term $P(b_t)=\max(0, b_t - b_g)$ increases the control variable proportionally to the distance from the setpoint, ii) the integral term $I(b_t) = \sum_{j=0}^t (b_j - b_g)$ corrects long-term deviations from the setpoint, and iii) the derivative term $D(b_t) = |b_{t} - b_{t-1}|$ uses the rate of change as a basic estimate of the future. The set point and the weights (also called gains) $k_P, k_I, k_D$ are hyperparameters. To compare against the strongest PID controller possible, we tuned these hyperparameters using multiple iterations of grid-search with exponential refinement per-patient, minimizing risk. Our final settings are presented in \textbf{Appendix \ref{app:pid_param}}.

\subsubsection{PID with Meal Announcements.}
This baseline, which is similar to available hybrid closed loop AP systems \citep{garg_glucose_2017, ruiz_effect_2012}, combines BB and PID into a control algorithm we call PID-MA. During meals, insulin boluses are calculated and applied as in the BB approach. The basal rate, instead of being fixed, is controlled by a PID algorithm, allowing for adaptation between meals. As above, we tune the gain parameters for the PID algorithm using sequential grid search to minimize risk.

\subsubsection{Oracle Architecture}
A deep RL approach to learning AP algorithms requires that the representation learned by the network contain sufficient information to control the system. As we are working with a simulator, we can explore the difficulty of this task in isolation, by replacing the observed state $\mathbf{s}_t$ with the ground-truth state $\mathbf{s}^*_t$. Though unavailable in real-world settings, this representation decouples the problem of learning a policy from that of inferring the state. Here, $Q_\theta$, $V_\psi$, and $\pi_\phi$ are fully connected with two hidden layers, each with 256 units. The network uses ReLU nonlinearities and BatchNorm \citep{ioffe2015batch}.

\subsection{Experimental Setup \& Evaluation}
To measure the utility of deep RL for the task of blood glucose control, we trained and tested our policies on data with different random seeds across 30 different simulated individuals. 
\paragraph{Training and Hyperparameters.}
We trained our models separately for each patient. They were trained from scratch for 300 epochs for RL-Scratch, and fine-tuned for 50 epochs for RL-Trans. They were trained with batch size 256 and an epoch length of 20 days. We used an experience replay buffer of size 1e6 and a discount factor of 0.99. We found that extensive training from-scratch was required to obtain consistent performance across test runs. We also found that too small of an epoch length could lead to dangerous control policies. We optimized the parameters of $Q_\theta$, $V_\psi$ and $\pi_\phi$ using Adam with a learning rate of $3E-4$. All deep networks were composed of two layers of GRU cells with a hidden state size of 128, followed by a fully-connected output layer. All network hyperparameters, including number and size of layers, were optimized on training seeds on a subset of the simulated patients for computational efficiency. Our networks were initialized using PyTorch defaults.

\paragraph{Evaluation.}
We measured the performance of $\pi_\phi$ on 10 days of validation data after each training epoch. After training, we evaluated on test data using the model parameters from the best epoch as determined by the validation data. While this form of model selection is not typical for RL, we found it led to significant changes in performance (see \textbf{Section \ref{sec:val_selection}}). Our model selection procedure first filters out runs that could not control blood glucose within safe levels over the validation run (glucose between 30-1000 mg/dL), then selects the epoch that achieved the lowest risk. We tested each patient-specific model on 1000 days of test data, broken into 100 independent 10 day rollouts. We trained and evaluated each approach 3 times, resulting in 3000 days of evaluation per method per person. 

We evaluated approaches using i) risk, the average Magni risk calculated over the 10-day test rollout, ii) \% time spent euglycemic (blood glucose levels between 70-180 mg/dL), iii) \% time hypo/hyperglycemic (blood glucose lower than 70 mg/dL or higher than 180 mg/dL respectively), and iv) \% of rollouts that resulted in a catastrophic failure, which we define as a run that achieves a minimal blood glucose level below 5 mg/dL (at which point recovery becomes highly unlikely). Note that while catastrophic failures are a major concern, our simulation process does not consider consuming carbohydrates in reaction to low blood glucose levels. This is a common strategy to avoid dangerous hypoglycemia in real life, and thus catastrophic failures, while serious, are manageable. The random seeds controlling noise, meals, and all other forms of randomness, were different between training, validation, and test runs. We test the statistical significance of differences between methods using Mood's median test for all metrics except for catastrophic failure rate, for which we use a binomial test.

\section{Experiments and Results}
Our experiments are divided into two broad categories: i) experiments showing the benefits of deep RL for blood glucose control relative to baseline approaches and ii) the challenges of using deep RL in this setting, and how to overcome them.

Throughout our experiments, we consider 3 variants of RL methods: i) RL-Scratch, our approach trained from scratch on every individual, ii) RL-Trans, which fine-tunes an RL-Scratch model from an arbitrary child/adolescent/adult, and iii) RL-MA, which uses RL-Scratch trained using the automated meal boluses from BB or PID-MA. We also report results on an Oracle approach, which is trained and evaluated using the ground truth simulator state.
 
 \begin{figure*}[!htbp]
    \centering
    \includegraphics[width=\linewidth]{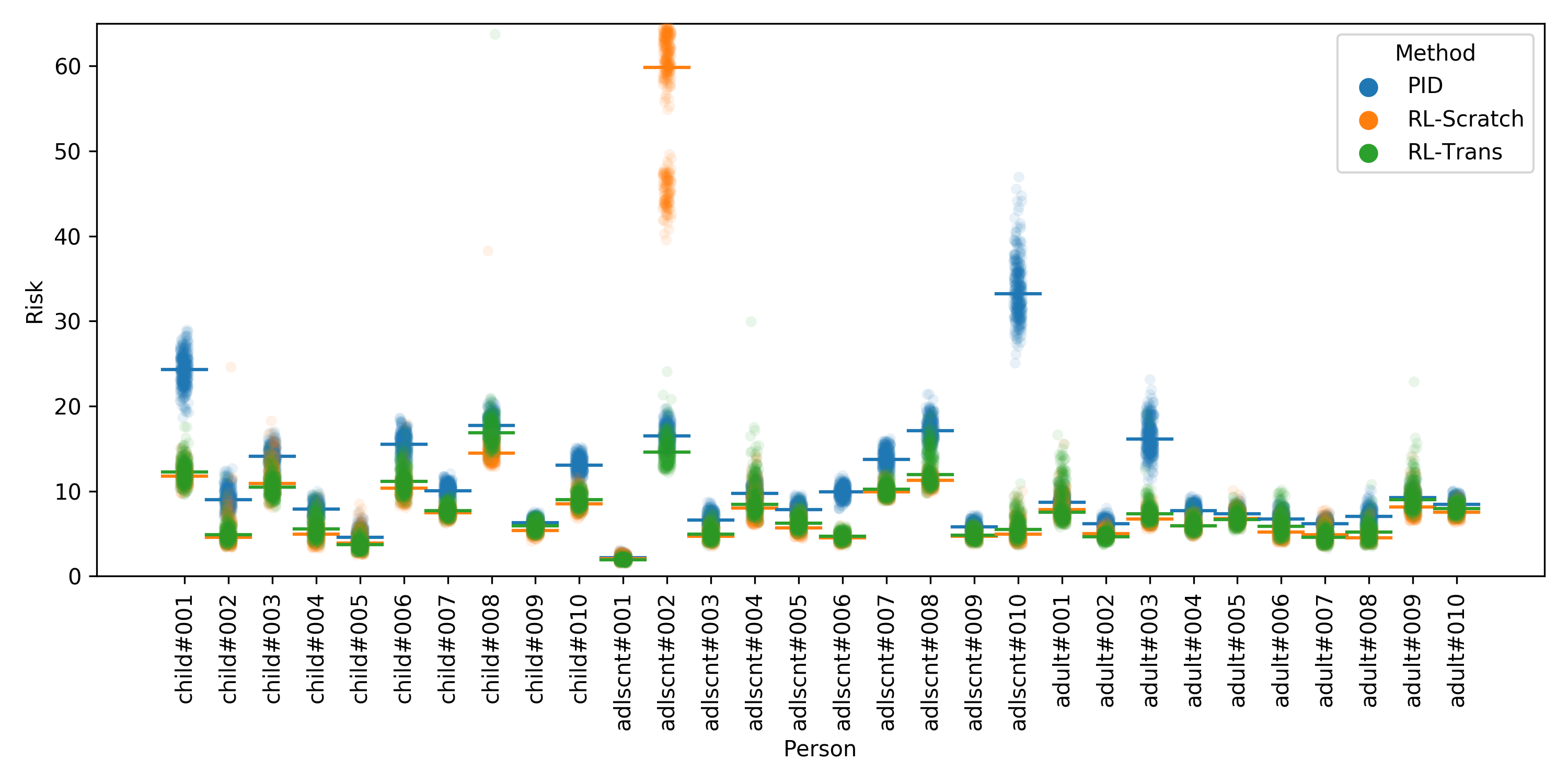}
    \caption{The risk over 10 days for different simulated patients using methods that do not require meal announcements. Each point corresponds to a different random test seed that controls the meal schedule and sensor noise, and the line indicates the median performance for each method on each patient. Results are presented across 3 random training seeds, controlling model initialization and randomness in training. We observe that, although there is a wide range in performance across and within individuals, The RL approaches tend to outperform PID.}
    \label{fig:full_risk_noma}
\end{figure*}

 \begin{figure*}[!htbp]
    \centering
    \includegraphics[width=\linewidth]{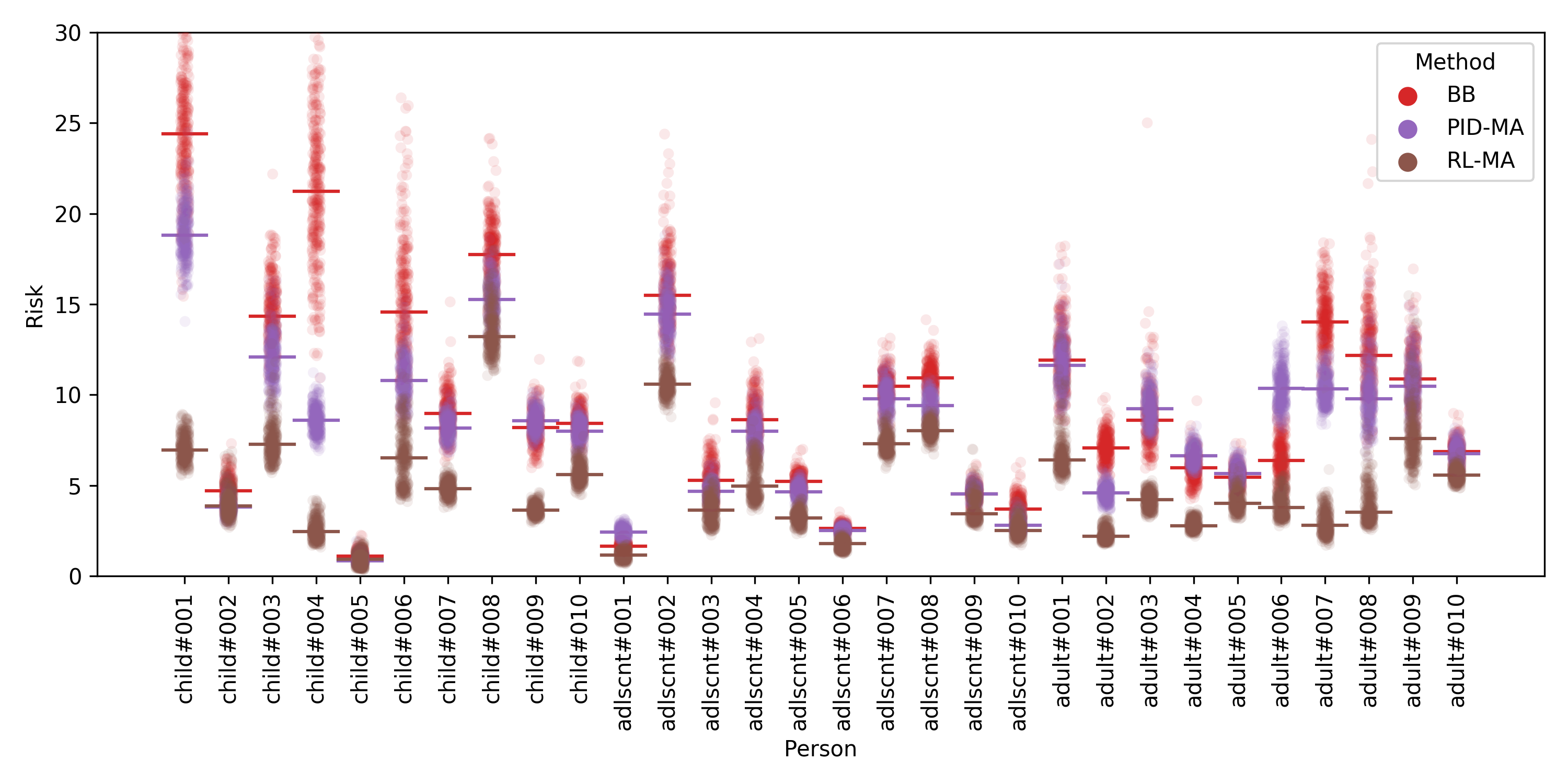}
    \caption{The risk over 10 days using methods that require meal announcements. PID-MA tends to outperform BB, and RL-MA outperforms PID-MA.}
    \label{fig:full_risk_ma}
\end{figure*}

\subsection{Advantages of Deep RL}\label{sec:advantages}
We compare our deep RL approaches to baselines with and without meal announcements across several metrics (\textbf{Section \ref{ssec:baseline}}). We then investigate two hypotheses for why deep RL is well suited to the problem of glucose management without meal announcements: 
\begin{itemize}
\item the high-capacity neural network, integral to the RL approaches, is able to quickly infer when meals occur (\textbf{Section \ref{ssec:meals}}), and 
\item the learning-based approach is able to adapt to predictable meal schedules better than a PID controller (\textbf{Section \ref{ssec:behavior}}). 
\end{itemize}

\subsubsection{Deep RL vs. Baseline Approaches}\label{ssec:baseline}
A comparison between the PID baseline to the RL approaches is presented in \textbf{Figure \ref{fig:full_risk_noma}}. Each point represents a different test rollout by a policy. For the RL approaches, the performance of each method is reported as the combination of 3 random training restarts. Among the 3 methods that do not require meal announcements, RL-Scratch performs best across patients (average rank 1.33), followed by RL-Trans (average rank 1.7), then PID (average rank 2.97). For each individual, we rank the 3 approaches in terms of median risk. We calculate average rank by taking the mean of each approach's rankings across all 30 individuals. Note that RL-Scratch, while achieving strong performance overall, reliably performs poorly on adolescent\#002. We discuss this issue in \textbf{Appendix \ref{app:ao2}}.

One major advantage of our proposed approach is its ability to achieve strong performance without meal announcements. This does not mean that it does not benefit from meal announcements, as shown in \textbf{Figure \ref{fig:full_risk_ma}}. Among the 3 methods that require meal announcements, RL-MA performs best (average rank 1.07), followed by PID-MA (average rank 2.13) then BB (average rank 2.8). 

We examine additional metrics in the results presented in \textbf{Table \ref{tab:risk}}. The difference between results that are bold, or bold and underlined, and the next best non-bold result (excluding RL-Oracle) are statistically significant with $p<0.001$. We observe that RL-MA equals or surpasses the performance of all non-oracle methods on all metrics, except for \% time spent hyperglycemia. Interestingly, all RL variants achieve lower median risk than PID-MA, which requires meal announcements. This is because the RL approaches achieve low levels of hypoglycemia, which the risk metric heavily penalizes (see \textbf{Figure \ref{fig:risk}}). Note that all methods, including PID-MA, were optimized to minimize this metric. Across patients, the RL methods achieve approximately 60-80\% time euglycemic, compared with $52\% \pm 19.6$\% observed in real human control \citep{ayanotakahara_carbohydrate_2015}. These results suggest that deep RL could be a valuable tool for closed-loop or hybrid closed-loop AP control.

\begin{table*}[!htbp]
    \centering
    \caption{Median risk, percent of time Eu/Hypo/Hyperglycemic, and failure rate calculated using 1000 days of simulation broken into 100 independent 10-day rollouts for each of 3 training seeds for 30 patients, totaling 90k days of evaluation (with interquartile range). Lower Magni Risk, Hypoglycemia, and Hyperglycemia are better, higher Euglycemia is better. Hybrid and Non-closed loop approaches (requiring meal announcements) are indicated with *. Approaches requiring a fully observed simulator state are indicated with $\dagger$. The non-oracle approach with the best average score is in bold and underlined, the best approach that does not require meal announcements is in bold.}
    \scalebox{0.83}{
    \begin{tabular}{lcccccc}
        \toprule
         & & Risk & Euglycemia & Hypoglycemia & Hyperglycemia & Failure \\
         & & $\downarrow$ & (\%) $\uparrow$ & (\%) $\downarrow$ & (\%) $\downarrow$ & (\%)  $\downarrow$ \\
        \midrule
        \multirow[c]{3}{*}{\rotatebox[origin=c]{90}{No MA}} & PID & 8.86 (6.8-14.3) & 71.68 (65.9-75.9) & 1.98 (0.3-5.5) & \textbf{24.71 (21.1-28.6)} & 0.12 \\
        & RL-Scratch & \textbf{6.50 (4.8-9.3)} & \textbf{72.68 (67.7-76.2)} & \textbf{0.73 (0.0-1.8)} & 26.17 (23.1-30.6) & \textbf{0.07} \\
        & RL-Trans & 6.83 (5.1-9.7) & 71.91 (66.6-76.2) & 1.04 (0.0-2.5) & 26.60 (22.7-31.0) & 0.22 \\
        \hline
        \multirow[c]{3}{*}{\rotatebox[origin=c]{90}{MA}} & BB$^*$ & 8.34 (5.3-12.5) & 71.09 (62.2-77.9) & 2.60 (0.0-7.2) & 23.85 (17.0-32.2) & 0.26 \\
        & PID-MA$^*$ & 8.31 (4.7-10.4) & 76.54 (70.5-82.0) & 3.23 (0.0-8.8) & \bf{\underline{18.74 (12.9-23.2)}} & \bf{\underline{0.00}} \\
        & RL-MA$^*$ & \bf{\underline{4.24 (3.0-6.5)}} & \bf{\underline{77.12 (71.8-83.0)}} & \bf{\underline{0.00 (0.0-0.9)}} & 22.36 (16.6-27.7) & \bf{\underline{0.00}} \\
        \midrule
        & RL-Oracle$^\dagger$ & 3.58 (1.9-5.4) & 78.78 (73.9-84.9) & 0.00 (0.0-0.0) & 21.22 (15.1-26.1) & 0.01 \\
        \bottomrule
    \end{tabular}
    }
    \label{tab:risk}
\end{table*}

\subsubsection{Detecting Latent Meals}\label{ssec:meals}
Our approach achieves strong blood glucose control without meal announcements, but how much of this is due to the ability to react to meals? To investigate this, we looked at the total proportion of insulin delivered on average after meals for PID and RL-Scratch, shown in \textbf{Figure \ref{fig:rl_advantage}a}. A method able to infer meals should use insulin rapidly after meals, as the sooner insulin is administered the faster glycemic spikes can be controlled while avoiding hypoglycemia. We observe that RL-Scratch administers the majority of its post-meal bolus within one hour of a meal, whereas PID requires over 90 minutes, suggesting RL-Scratch can indeed better infer meals. Interestingly, when considering the percentage of total daily insulin administered in the hour after meals, RL-Scratch responds even more aggressively than BB or PID-MA (54.7\% \textit{vs.} 48.5\% and 47.3\% respectively). This demonstrates that our RL approach is able to readily react to latent meals shortly after they have occurred. 

\subsubsection{Ability to Adapt to Predictable Meal Schedules}\label{ssec:behavior}
\begin{figure}
    \centering
    \begin{subfigure}{}
        \includegraphics[width=.4\linewidth]{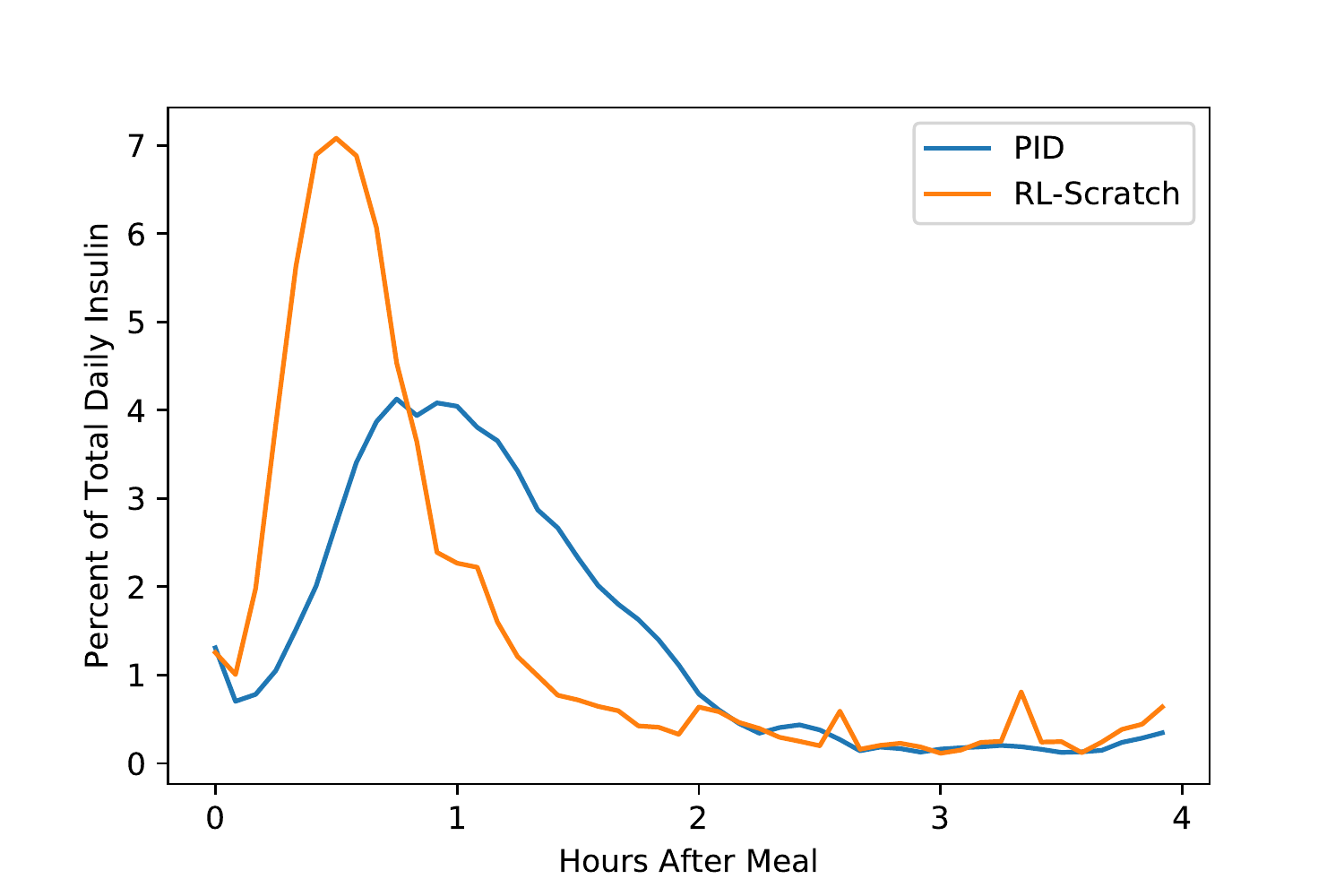}
    \end{subfigure}
    \begin{subfigure}{}
        \includegraphics[width=.4\linewidth]{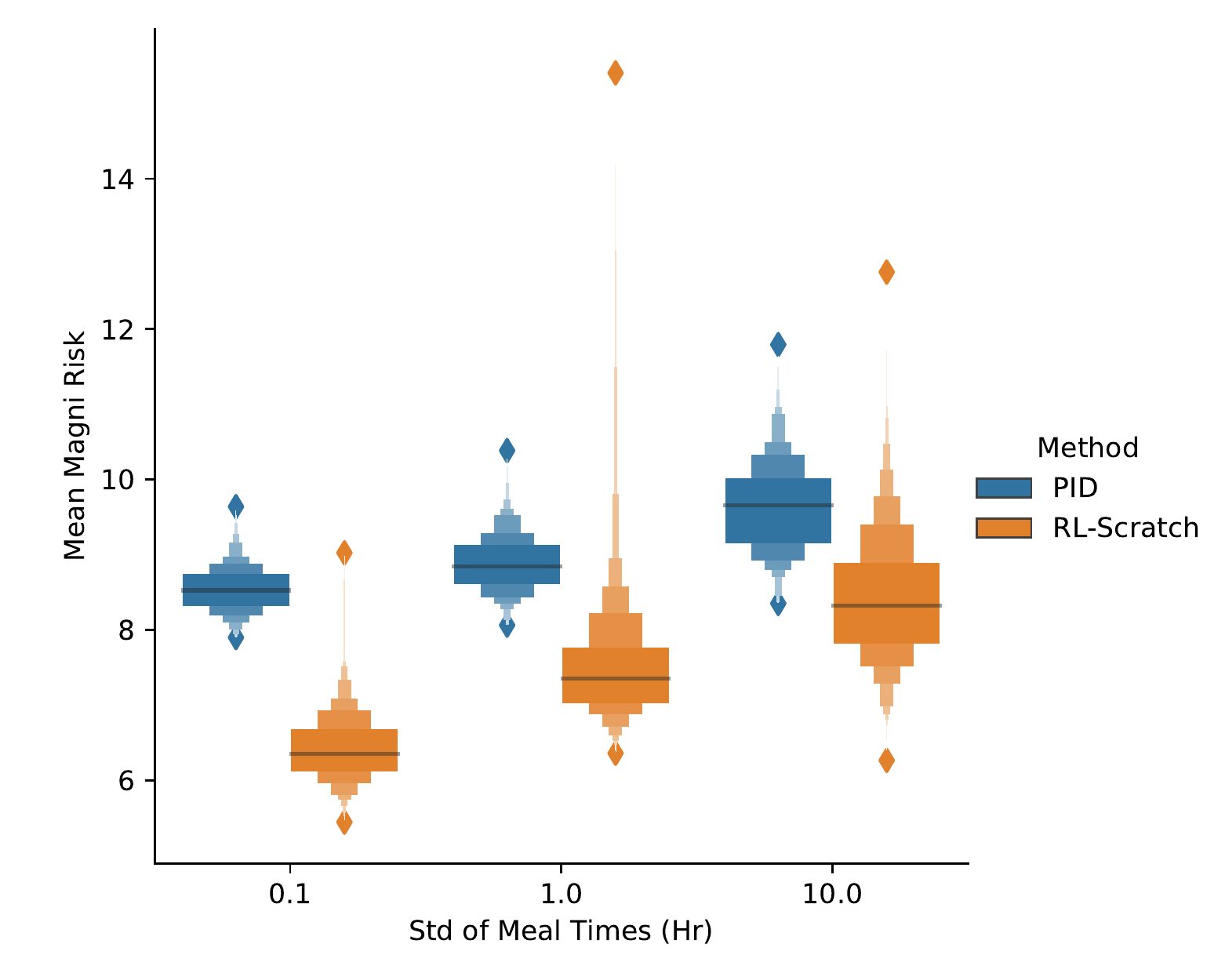}
    \end{subfigure}
    \caption{a) The average amount of insulin (in percent of total daily insulin) provided after meals for PID and RL-Scratch (note: RL-Trans, unshown, is very similar to RL-Scratch). RL-Scratch is able respond to meals more quickly than PID, with insulin peaking 30 minutes post-meal as opposed to roughly 60 minutes for PID. Additionally, the RL approach finishes delivering most post-meal insulin after 1hr, PID takes over 90 minutes. b) The distribution of average risk scores over 300 10-day rollouts for Adult 1 using meal schedules with varying amounts of predictability (meal time standard deviation). While PID performs better with more regularly spaced meals (median risk lowers from 9.66 at std=10 to 8.53 at std=0.1, a 12\% decrease), RL-Scratch sees a larger proportional and absolute improvement (median risk lowers from 8.33 at std=10 to 6.36 at std=0.1, a 24\% decrease).}\label{fig:rl_advantage}
\end{figure}

We hypothesize that one advantage of RL is its ability to compensate for predictable variations (such as meals) in the environment, improving control as the environment becomes more predictable. To test this benefit, we changed the meal schedule generation procedure outlined in \textbf{Algorithm \ref{alg:schedule}} (\textbf{Appendix \ref{app:meal}}) for Adult 1. We removed the small `snack' meals, set all meal occurrence probabilities to 1, and made meal amounts constant (\textit{i.e.}, each day Adult 1 consumes an identical set of meals). We then evaluated both the PID model and RL-Scratch on 3 variations of this environment, characterized by the standard deviation of the meal times (either 0.1, 1, or 10 hours). This tests the ability of each method to take advantage of patterns in the environment. As the standard deviation decreases, the task becomes easier for two reasons: i) there are fewer instances where two meals occur in quick succession, and ii) the meals become more predictable. The results are presented in \textbf{Figure \ref{fig:rl_advantage}b}. We observe that both methods improve in performance as the standard deviation decreases, likely due to (i). However, while RL-Scratch outperforms PID under all settings, the difference increases as the standard deviation of meal times decreases, suggesting RL is better able to leverage the predictability of meals. Specifically, mean risk decreases by roughly 12\% for the PID approach (from 9.65 to 8.54), whereas it decreases nearly 24\% for the RL approach (from 8.40 to 6.42). This supports our hypothesis that RL is better able to take advantage of predictable meal schedules.
\vspace{-3pt}

\subsection{Challenges for Deep RL}
While in the previous section we demonstrated several advantages of using deep RL for blood glucose control, here we emphasize that the application of deep RL to this task and its evaluation are non-trivial. Specifically, in this section we:

\begin{itemize}
\item demonstrate the importance of our action space formulation for performance (\textbf{Section \ref{ssec:actionspace}}), 
\item illustrate the critical need for careful and extensive validation, both for model selection and evaluation (\textbf{Section \ref{sec:val_selection}}). 
\item show that, applied naively, deep RL leads to an unacceptable catastrophic failure rate, and present three simple approaches to improve this (\textbf{Section \ref{ssec:failures}}),  
\item address the issue of sample complexity with simple policy transfer (\textbf{Section \ref{ssec:transfer}}).
\end{itemize}

\subsubsection{Developing an Effective Action Space}\label{ssec:actionspace}
One challenging element of blood glucose control in an RL setting is defining the action space. Insulin requirements vary significantly by person (from 16 to 60 daily units in the simulator population we use), and throughout most of the day, insulin requirements are much lower than after meals. To account for these challenges, we used a patient-specific action space, where much of the space corresponds to delivering no insulin (discussed in \textbf{Section \ref{sec:architecture}}). We perform an ablation study to test the impact of these decisions. On an arbitrarily chosen patient (child\#001), we shifted the $tanh$ output to remove the negative insulin space. This increased the catastrophic failure rate from 0\% (on this patient) to 6.6\%. On a challenging subset of 4 patients (indicated in \textbf{Appendix \ref{app:bb}}), we looked at the effect of removing the patient-specific action scaling $\omega_b$. This resulted in a 13\% increase in median risk from 9.92 to 11.23. These results demonstrate that a patient-specific action space that encourages conservative behavior can improve performance.

\subsubsection{Potential Pitfalls in Evaluation}\label{sec:val_selection}
In our experiments, we observed two key points for model evaluation: i) while often overlooked in RL, using validation data for model selection during training was key to achieving good performance, and ii) without evaluating on far more data than is typical, it was easy to underestimate the catastrophic failure rate.

\begin{figure}[!htbp]
    \centering
    \begin{subfigure}
        \centering
        \includegraphics[width=0.45\linewidth]{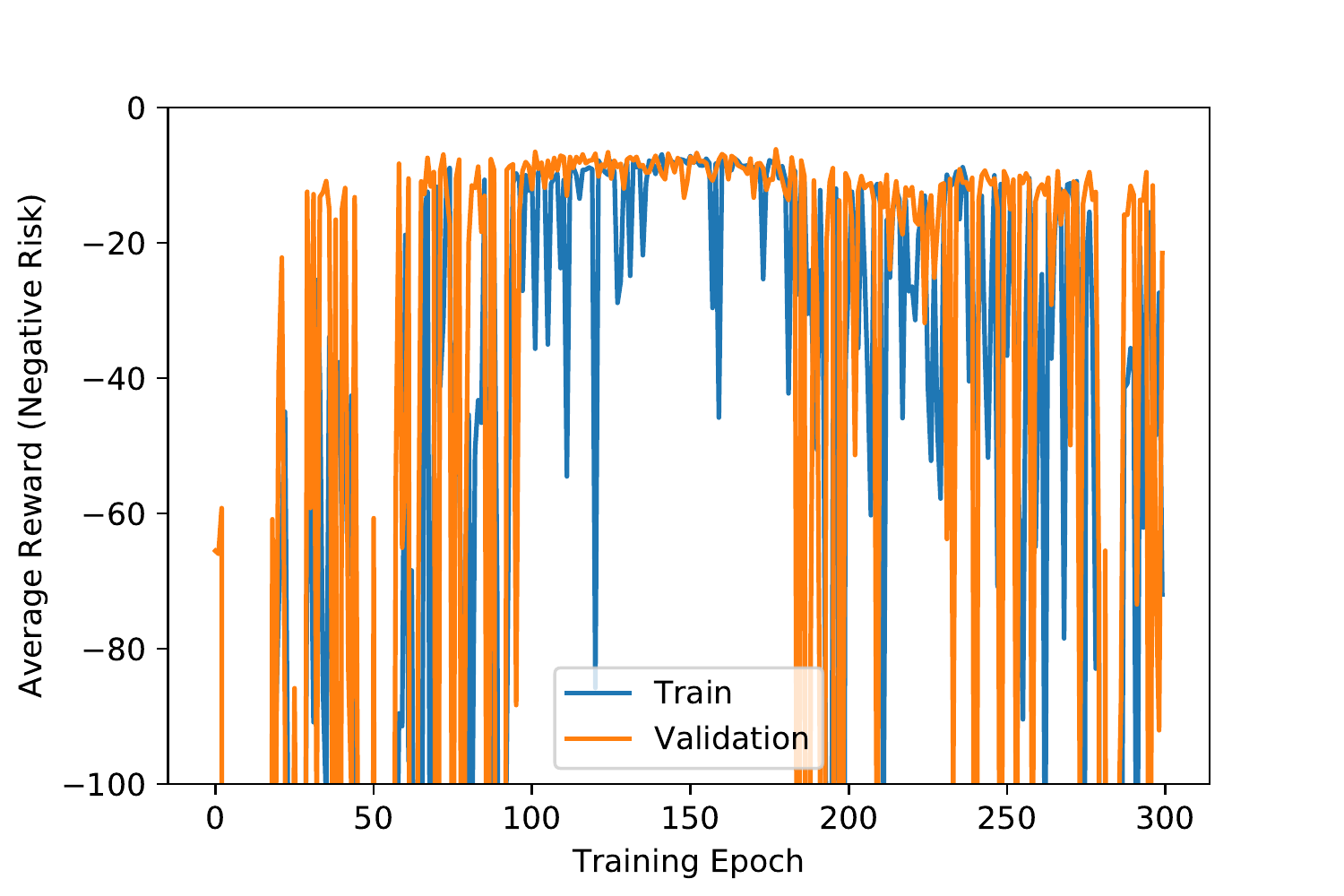}
        \label{fig:validation_curve}
    \end{subfigure}
    \begin{subfigure}{}
        \includegraphics[width=0.4\linewidth]{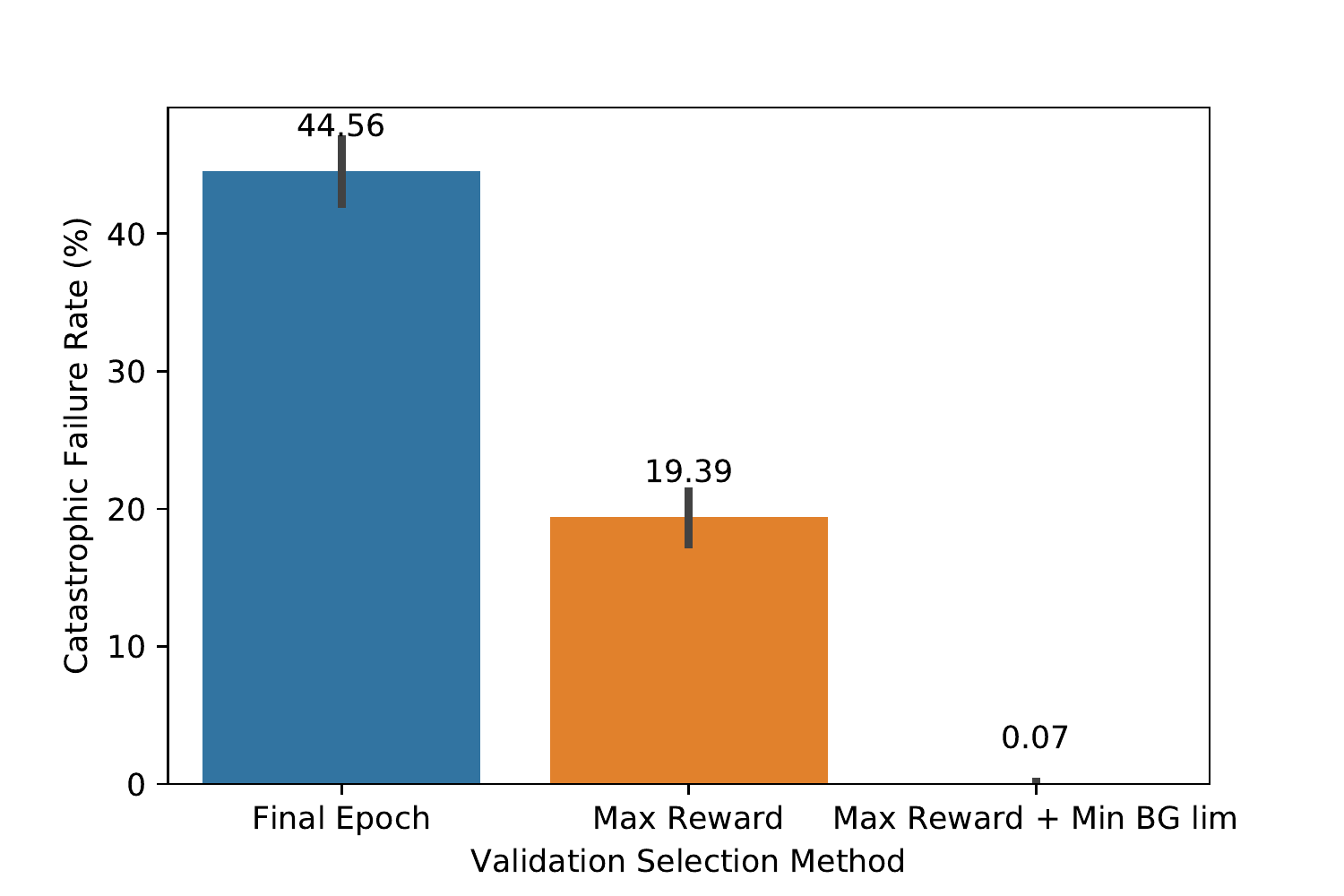}
        \label{fig:validation_perf}
    \end{subfigure}
    \caption{a) The training and validation curve (average reward) for adult\#009. Note the periods of instability affect both training and validation performance. b) Catastrophic failure rate over all patients for 3 methods of model selection: i) selecting the final training epoch, ii) selecting the epoch that achieved minimal risk, and iii) selecting the minimal risk epoch that maintained blood glucose above 30 mg/dL. We see large differences in performance depending on the model selection strategy.}\label{fig:validation_selection}
\end{figure}

\paragraph{Performance instability necessitates careful model selection.}
Off-policy RL with function approximation, particularly deep neural networks, is known to be unstable \citep{sutton_reinforcement_2018, gottesman2018evaluating}. As a result, we found it was extremely important to be careful in selecting which network (and therefore policy) to evaluate. In \textbf{Figure \ref{fig:validation_selection}a}, we show the fluctuation in validation performance over training for Adult\#009. While performance increases on average over the course of training (at least initially), there are several points where performance degrades considerably. \textbf{Figure \ref{fig:validation_selection}b} shows how performance averaged over all patients changes depending on the approach used to select the policy for evaluation. When we simply evaluate using the final epoch, almost half of test rollouts end in a catastrophic failure. Surprisingly, even when we select the model that minimized risk on the validation set, nearly a fifth of rollouts fail. However, by first limiting our pool of models to those that achieve a minimum blood glucose level of at least 30 mg/dL over the validation data, we reduce the catastrophic failure rate to $0.07\%$. As performance instability has been noted in other RL domains \citep{islam_reproducibility_2017, henderson_deep_2018}, this observation is likely relevant to other applications of deep RL in healthcare.

\paragraph{Extensive evaluation is necessary.} 
Health applications are often safety critical. Thus, the susceptibility of deep RL to unexpected or unsafe behavior can pose a significant risk \citep{leike2017ai, futoma_identifying_2020}. While ongoing work seeks to provide safety guarantees in control applications using deep learning \citep{achiam_constrained_2017, futoma_identifying_2020}, it is important that practitioners take every step to evaluate the safety of their approaches. While it is typical to evaluate RL algorithms on a small number of rollouts \citep{henderson_deep_2018}, in our work we saw how easy it can be to miss unsafe behavior, even with significant testing. We examined the number of catastrophic failures that occurred using RL-Trans using different evaluation sets. Over our full evaluation set of 9,000 10-day rollouts, we observed 20 catastrophic failures (a rate of 0.22\%). However, when we only evaluated using the first 5 test seeds, which is still over 12 years of data, we observed 0 catastrophic failures. Additionally, when we evaluated using 3-day test rollouts instead of 10, we only observed only 5 catastrophic failures (a rate of .05\%), suggesting that longer rollouts result in a higher catastrophic failure rate. These results demonstrate that, particularly when dealing with noisy observations, it is critical to evaluate potential policies using a large number of different, lengthy rollouts.

\subsubsection{Reducing Catastrophic Failures}\label{ssec:failures}
Due to their potential danger, avoiding catastrophic failures was a significant goal of this work. The most direct approach we used was to modify the reward function, using a large termination penalty to discourage dangerous behavior. While unnecessary for fine-tuning policies, when training from scratch this technique was crucial. On a subset of 6 patients (see \textbf{Appendix \ref{app:patient}}), including the termination penalty reduced the catastrophic failure rate from $4.2\%$ to $0.2\%$.

We found varying the training data also had a major impact on the catastrophic failure rate. During training, every time we reset the environment we used a different random seed (which controls meal schedule and sensor noise). Note that the pool of training, validation, and test seeds were non-overlapping. On a challenging subset of 7 patients (described in \textbf{Appendix \ref{app:patient}}), we ran RL-Scratch with and without this strategy. The run that varied the training seeds had a catastrophic failure rate of $0.15\%$, the run that didn't had a $2.05\%$ rate (largely driven by adult\#009, who reliably had the highest catastrophic failure rate across experiments).

Other approaches can further improve stability. In \textbf{Table \ref{tab:risk}}, our RL results are averaged over three random restarts in training. This was done to demonstrate that our learning framework is robust to randomness in training data and model initialization. However, in a real application it would make more sense to select (using validation data) one model for use out of the random restarts. We apply this approach in \textbf{Table \ref{tab:ss_risk}}, choosing the seed that obtained the best performance according to our model selection criteria. This improves all the RL methods. Most notably, it further reduces the catastrophic failure rate for the approaches without meal announcements (0.07\% to 0\% for RL-Scratch, and 0.22\% to 0.13\% for RL-Trans). 

\begin{table*}[t!]
    \centering
    \caption{Risk and percent of time Eu/Hypo/Hyperglycemic calculated for the RL approaches treating the 3 training seeds as different random restarts. The stability of the Scratch and Trans approaches improves relative to performance in Table \ref{tab:risk}.}
    \scalebox{0.89}{
    \begin{tabular}{lccccc}
        \toprule
         & Risk & Euglycemia & Hypoglycemia & Hyperglycemia & Failure Rate \\
         & $\downarrow$ & (\%) $\uparrow$ & (\%) $\downarrow$ & (\%) $\downarrow$ & (\%)  $\downarrow$ \\
        \midrule
        RL-Scratch & 6.39 (4.7-8.9) & 72.96 (69.1-76.6) & 0.62 (0.0-1.7) & 25.96 (22.7-29.6) & 0.00 \\
        RL-Trans & 6.57 (5.0-9.3) & 71.56 (67.0-75.7) & 0.80 (0.0-1.9) & 27.19 (23.4-31.2) & 0.13 \\
        RL-MA & 3.71 (2.7-6.3) & 77.36 (72.7-83.2) & 0.00 (0.0-0.5) & 22.45 (16.7-26.9) & 0.00 \\
        \bottomrule
    \end{tabular}
    }
    \label{tab:ss_risk}
\end{table*}

\subsubsection{Sample Efficiency and Policy Transfer}\label{ssec:transfer}
While RL-Scratch achieves strong performance on average, it requires a large amount of patient-specific data: 16.5 years per patient. While RL-Trans reduced this amount, it still required over 2 years of patient-specific data, which for most health applications would be infeasible. Thus, we investigated how performance degrades as less data is used.

\begin{figure}
    \centering
    \includegraphics[width=.5\linewidth]{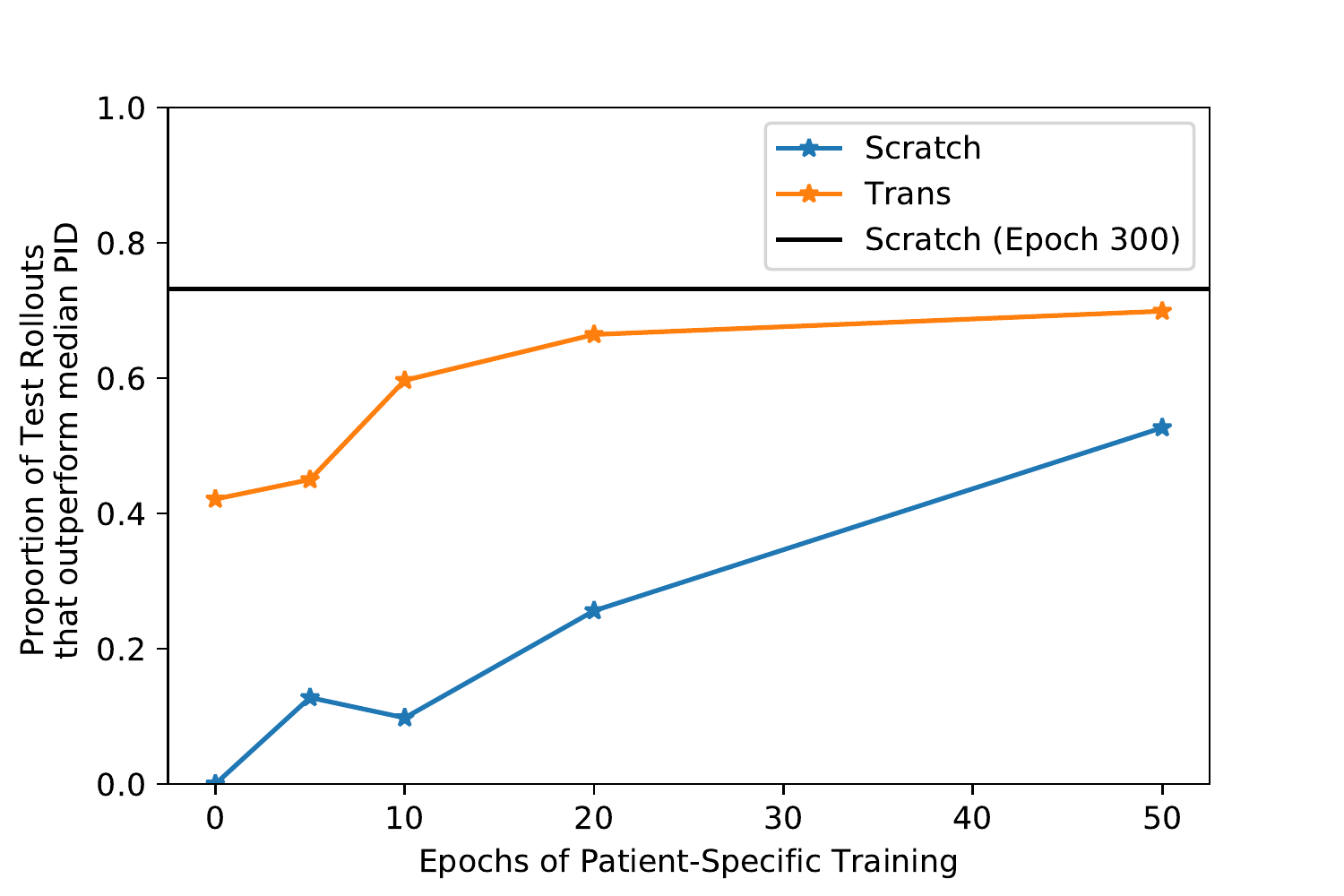}
    \caption{The proportion of test rollouts where RL-Scratch and RL-Trans outperform the median PID risk with different amounts of patient-specific training. We see that without any patient-specific data RL-Trans performs better than PID in 40\% of rollouts. RL-Scratch requires a significant amount of patient-specific data before achieving comparable performance.}\label{fig:finetune}
\end{figure}

In \textbf{Figure \ref{fig:finetune}}, we show the average policy performance by epoch for both RL-Scratch and RL-Trans relative to the PID controller. Note the epoch determines the maximum possible epoch for our model selection, not the actual chosen epoch. We see that far less training is required to achieve good performance with RL-Trans. In over 40\% of rollouts, RL-Trans outperforms PID with no patient-specific data (median risk 10.31), and with 10 epochs of training (roughly half a year of data) RL-Trans outperforms PID in the majority of rollouts (59.6\%; median risk 7.95). 

Interestingly, the lack of patient-specific data does not appear to cause excessive catastrophic failures. With no patient-specific data the failure rate is 0\%, after 5 epochs of training it has risen to .5\%, and then declines over training to the final value of .22\%. This implies two things: i) patient-specific training can increase the catastrophic failure rate, possibly by learning overly aggressive treatment policies, and ii) our current model selection procedure does not minimize the catastrophic failure rate. We do not observe this for RL-Scratch, where all epochs under 50 achieve a catastrophic error rate of over 17\%. These results suggest that our simple transfer approach can be effective even with limited amounts of patient-specific data.

\section{Conclusion}
In this work, we demonstrated how deep RL can lead to significant improvements over alternative approaches to blood glucose control, with and without meal announcements (\textbf{Section \ref{ssec:baseline}}). We provide insight into why (\textbf{Section \ref{ssec:meals}}) and when (\textbf{Section \ref{ssec:behavior}}) we would expect to see RL perform better. We demonstrated the importance of a robust action space in patient-specific tasks (\textbf{Section \ref{ssec:actionspace}}), showed how careful and extensive validation is necessary for realistic evaluation of performance(\textbf{Section \ref{sec:val_selection}}), investigated several simple and general approaches to improving the stability of deep RL (\textbf{Section \ref{ssec:failures}}), and developed a simple approach to reduce the requirements of patient-specific data (\textbf{Section \ref{ssec:transfer}}). While the main goal of this paper was to advance a clinical application, the challenges we encountered and the solutions they inspired are likely to be of interest to researchers applying RL to healthcare more broadly. 

While our results in applying deep RL to blood glucose control are encouraging, they come with several limitations. First, our results are based on simulation. The simulator may not adequately capture variation across patients or changes in the glucoregulatory system over time. In particular, the virtual patient population the simulator comes equipped with does not differentiate individuals based on demographic information, such as gender and ethnicity. Thus, the applicability of our proposed techniques to all relevant patient populations cannot be assessed. However, as an FDA-approved substitute for animal trials \citep{kovatchev_silico_2009}, success in using this simulator is a nontrivial accomplishment. Second, we define a reward function based on risk. Though optimizing this risk function should lead to tight glucose control, it could lead to excess insulin utilization (as its use is unpenalized). Future work could consider resource-aware variants of this reward. Third, our choice of a 4-hour state space discourages learning long-term patterns or trends. In our environment, this did not reduce performance relative to a longer input history, but this could be important for managing blood glucose levels in more realistic simulators or real-world cases \cite{visentin_uva/padova_2018}. Finally, we emphasize that blood glucose control is a safety-critical application. An incorrect dose of insulin could lead to life-threatening situations. Many of the methods we examined, even those that achieve good average performance, are susceptible to catastrophic failures. We have investigated several ways to minimize such failures, including modifying the reward function and selecting models across multiple random restarts. While the results from \textbf{Table \ref{tab:ss_risk}} suggest these approaches mitigate catastrophic failures, the results of \textbf{Section \ref{sec:val_selection}} show such empirical evaluations can miss failure cases. To enable researchers to better explore and correct these limitations, we evaluated on an open-source simulator and made all the code required to reproduce our experiments publicly available.

\acks{This work was supported by JDRF. The views and conclusions in this document are those of the authors and should not be interpreted as necessarily representing the official policies, either expressed or implied of JDRF.}

\bibliography{references}

\clearpage

\appendix
\section{Appendix}
\subsection{Harrison-Benedict Meal Generation Algorithm}
\label{app:meal}
See \textbf{Algorithm \ref{alg:schedule}}. In short, this meal schedule calculates BMR for each simulated individual using the Harrison-Benedict equation \citep{harris_biometric_1919}. This is used to estimate expected daily carbohydrate consumption, assuming individuals eat a reasonably low-carb diet where 45\% of calories come from carbohydrates. Daily carbohydrates were divided between 6 potential meals: breakfast, lunch, dinner, and 3 snacks. The occurrence probability of the meal and expected size of the meal was set such that the expected number of carbs eaten per day matched the BMR-derived estimate. Note, in our experiments our implementation of of this meal schedule incorrectly estimated ages for many patients, particularly adults. This effect was fairly minor, resulting in no more than a 10\% change in estimated BMR.

\begin{algorithm}[!htbp]
   \caption{Generate Meal Schedule }
   \label{alg:schedule}
\begin{algorithmic}
   \STATE {\bfseries Input:} body weight $w$, age $a$, height $h$, number of days $n$
   \STATE $BMR = 66.5 + (13.75 * w) + (5.003 * h) - (6.755 * a)$
   \STATE $ExpectedCarbs = (BMR*0.45)/4$ \COMMENT{45\% of calories assumed from carbs, 4 calories per carb}
   \STATE $MealOcc = [0.95, 0.3, 0.95, 0.3, 0.95, 0.3]$
   \STATE $TimeLowerBounds = [5, 9, 10, 14, 16, 20]*12$
   \STATE $TimeUpperBounds = [9, 10, 14, 16, 20, 23]*12$
   \STATE $TimeMean = [7, 9.5, 12, 15, 18, 21.5]*12$
   \STATE $TimeStd = [1, .5, 1, .5, 1, .5]*12$
   \STATE $AmountMean = [0.250, 0.035, 0.295, 0.035, 0.352, 0.035] * ExpectedCarbs *1.2$
   \STATE $AmountStd = AmountMean * 0.15$
   \STATE $Days = []$
   \FOR{$i \in [1, \dots, n]$}
   \STATE $M = [0]_{j=1}^{288}$
   \FOR{$j \in [1, \dots, 6]$}
   \STATE $m \sim Binomial(MealOcc[j])$
   \STATE $lb = TimeLowerBounds[j]$
   \STATE $ub = TimeUpperBounds[j]$
   \STATE $\mu_t = TimeMean[j]$
   \STATE $\sigma_t = TimeStd[j]$
   \STATE $\mu_a = AmountMean[j]$
   \STATE $\sigma_a = AmountStd[j]$
   \IF{ $m$}
   \STATE $t \sim Round(TruncNormal(\mu_t, \sigma_t, lb, ub))$
   \STATE $c \sim Round(max(0, Normal(\mu_a, \sigma_a)))$
   \STATE $M[t] = c$
   \ENDIF
   \ENDFOR
   \STATE $Days.append(M)$
   \ENDFOR
\end{algorithmic}
\end{algorithm}

\clearpage

\subsection{BB Parameters}
\label{app:bb}
The parameters are presented in \textbf{Table \ref{tab:bb}}. The simulator provides age and TDI for each individual. To calculate CR and CF we use equations $CR = 500/TDI$ and $CF = 1800/TDI$, which were provided to us via a clinical consultation. The resulting CR and CF we used differed from those provided in the baseline Simglucose download from \cite{xie_simglucose_2018}, in practice we found our values led to better performance. 

\begin{table}
    \centering
    \begin{tabular}{lcccc}
    Person & CR & CF & Age & TDI \\
    \hline
    child\#001 & 28.62 & 103.02 & 9 & 17.47 \\
    child\#002 & 27.51 & 99.02 & 9 & 18.18 \\
    child\#003 & 31.21 & 112.35 & 8 & 16.02 \\
    child\#004 & 25.23 & 90.84 & 12 & 19.82 \\
    child\#005 & 12.21 & 43.97 & 10 & 40.93 \\
    child\#006 & 24.72 & 89.00 & 8 & 20.22 \\
    child\#007 & 13.81 & 49.71 & 9 & 36.21 \\
    child\#008 & 23.26 & 83.74 & 10 & 21.49 \\
    child\#009 & 28.75 & 103.48 & 7 & 17.39 \\
    child\#010 & 24.21 & 87.16 & 12 & 20.65 \\
    adolescent\#001 & 13.61 & 49.00 & 18 & 36.73 \\
    adolescent\#002 & 8.06 & 29.02 & 19 & 62.03 \\
    adolescent\#003 & 20.62 & 74.25 & 15 & 24.24 \\
    adolescent\#004 & 14.18 & 51.06 & 17 & 35.25 \\
    adolescent\#005 & 14.70 & 52.93 & 16 & 34.00 \\
    adolescent\#006 & 10.08 & 36.30 & 14 & 49.58 \\
    adolescent\#007 & 11.46 & 41.25 & 16 & 43.64 \\
    adolescent\#008 & 7.89 & 28.40 & 14 & 63.39 \\
    adolescent\#009 & 20.77 & 74.76 & 19 & 24.08 \\
    adolescent\#010 & 15.07 & 54.26 & 17 & 33.17 \\
    adult\#001 & 9.92 & 35.70 & 61 & 50.42 \\
    adult\#002 & 8.64 & 31.10 & 65 & 57.87 \\
    adult\#003 & 8.86 & 31.90 & 27 & 56.43 \\
    adult\#004 & 14.79 & 53.24 & 66 & 33.81 \\
    adult\#005 & 7.32 & 26.35 & 52 & 68.32 \\
    adult\#006 & 8.14 & 29.32 & 26 & 61.39 \\
    adult\#007 & 11.90 & 42.85 & 35 & 42.01 \\
    adult\#008 & 11.69 & 42.08 & 48 & 42.78 \\
    adult\#009 & 7.44 & 26.78 & 68 & 67.21 \\
    adult\#010 & 7.76 & 27.93 & 68 & 64.45 \\
    \end{tabular}
    \caption{Basal-Bolus Parameters}
    \label{tab:bb}
\end{table}

\clearpage

\subsection{PID and PID-MA parameters}
\label{app:pid_param}
\begin{table}[!htbp]
    \centering
    \begin{tabular}{lccc}
     & $k_p$ & $k_i$ & $k_d$ \\
     \hline
    child\#001 & -3.49E-05 & -1.00E-07 & -1.00E-03 \\
    child\#002 & -3.98E-05 & -2.87E-08 & -3.98E-03 \\
    child\#003 & -6.31E-05 & -1.74E-08 & -1.00E-03 \\
    child\#004 & -6.31E-05 & -1.00E-07 & -1.00E-03 \\
    child\#005 & -1.00E-04 & -2.87E-08 & -6.31E-03 \\
    child\#006 & -3.49E-05 & -1.00E-07 & -1.00E-03 \\
    child\#007 & -3.98E-05 & -6.07E-08 & -2.51E-03 \\
    child\#008 & -3.49E-05 & -3.68E-08 & -1.00E-03 \\
    child\#009 & -3.49E-05 & -1.00E-07 & -1.00E-03 \\
    child\#010 & -4.54E-06 & -3.68E-08 & -2.51E-03 \\
    adolescent\#001 & -1.74E-04 & -1.00E-07 & -1.00E-02 \\
    adolescent\#002 & -1.00E-04 & -1.00E-07 & -6.31E-03 \\
    adolescent\#003 & -1.00E-04 & -1.00E-07 & -3.98E-03 \\
    adolescent\#004 & -1.00E-04 & -1.00E-07 & -4.79E-03 \\
    adolescent\#005 & -6.31E-05 & -1.00E-07 & -6.31E-03 \\
    adolescent\#006 & -4.54E-10 & -1.58E-11 & -1.00E-02 \\
    adolescent\#007 & -1.07E-07 & -6.07E-08 & -6.31E-03 \\
    adolescent\#008 & -4.54E-10 & -4.54E-12 & -1.00E-02 \\
    adolescent\#009 & -6.31E-05 & -1.00E-07 & -3.98E-03 \\
    adolescent\#010 & -4.54E-10 & -4.54E-12 & -1.00E-02 \\
    adult\#001 & -1.58E-04 & -1.00E-07 & -1.00E-02 \\
    adult\#002 & -3.98E-04 & -1.00E-07 & -1.00E-02 \\
    adult\#003 & -4.54E-10 & -1.00E-07 & -1.00E-02 \\
    adult\#004 & -1.00E-04 & -1.00E-07 & -3.98E-03 \\
    adult\#005 & -3.02E-04 & -1.00E-07 & -1.00E-02 \\
    adult\#006 & -2.51E-04 & -2.51E-07 & -1.00E-02 \\
    adult\#007 & -1.22E-04 & -3.49E-07 & -2.87E-03 \\
    adult\#008 & -1.00E-04 & -1.00E-07 & -1.00E-02 \\
    adult\#009 & -1.00E-04 & -1.00E-07 & -1.00E-02 \\
    adult\#010 & -1.00E-04 & -1.00E-07 & -1.00E-02 \\
    \end{tabular}
    \caption{PID parameters}
    \label{tab:pid}
\end{table}

\begin{table}[!htbp]
    \centering
    \begin{tabular}{lccc}
     & $k_p$ & $k_i$ & $k_d$ \\
     \hline
    child\#001 & -5.53E-09 & -1.00E-07 & -3.49E-04 \\
    child\#002 & -1.00E-04 & -2.87E-08 & -1.00E-03 \\
    child\#003 & -1.00E-05 & -2.87E-08 & -1.00E-03 \\
    child\#004 & -6.31E-05 & -1.00E-07 & -1.00E-03 \\
    child\#005 & -1.00E-04 & -1.00E-07 & -3.31E-03 \\
    child\#006 & -1.00E-05 & -3.68E-08 & -1.00E-03 \\
    child\#007 & -2.35E-07 & -1.00E-07 & -1.00E-03 \\
    child\#008 & -4.72E-06 & -2.87E-08 & -1.00E-03 \\
    child\#009 & -1.00E-05 & -1.00E-07 & -3.49E-04 \\
    child\#010 & -3.49E-05 & -4.72E-08 & -1.00E-03 \\
    adolescent\#001 & -1.00E-04 & -4.72E-08 & -6.31E-03 \\
    adolescent\#002 & -1.00E-05 & -1.00E-07 & -3.49E-03 \\
    adolescent\#003 & -6.31E-05 & -1.00E-07 & -2.09E-03 \\
    adolescent\#004 & -6.31E-05 & -1.00E-07 & -2.51E-03 \\
    adolescent\#005 & -4.79E-05 & -1.00E-07 & -3.98E-03 \\
    adolescent\#006 & -1.00E-04 & -1.00E-07 & -2.75E-03 \\
    adolescent\#007 & -1.00E-05 & -1.00E-07 & -3.02E-03 \\
    adolescent\#008 & -1.58E-09 & -1.00E-07 & -2.75E-03 \\
    adolescent\#009 & -3.98E-05 & -1.00E-07 & -1.91E-03 \\
    adolescent\#010 & -1.00E-04 & -1.00E-07 & -4.37E-03 \\
    adult\#001 & -1.07E-07 & -1.00E-07 & -4.37E-03 \\
    adult\#002 & -1.58E-04 & -1.00E-07 & -4.37E-03 \\
    adult\#003 & -7.59E-05 & -1.00E-07 & -2.51E-03 \\
    adult\#004 & -1.00E-04 & -1.00E-07 & -1.00E-03 \\
    adult\#005 & -1.07E-07 & -1.00E-07 & -6.31E-03 \\
    adult\#006 & -1.00E-04 & -1.00E-07 & -1.00E-02 \\
    adult\#007 & -1.58E-04 & -2.51E-07 & -3.02E-03 \\
    adult\#008 & -1.58E-05 & -1.00E-07 & -3.98E-03 \\
    adult\#009 & -4.54E-10 & -1.00E-07 & -6.31E-03 \\
    adult\#010 & -3.98E-05 & -1.00E-07 & -4.37E-03 \\
    \end{tabular}
    \caption{PID-MA parameters}
    \label{tab:pid_ma}
\end{table}

\clearpage

\subsection{Relevant Patient Subgroups}\label{app:patient}
For tuning our models and selecting hyperparameters, we focused on one challenging but representative individual from each category. In particular, we used child\#001, adolescent\#004, and adult\#001.

For our action space ablation, we examined the subset of 4 individuals who were most prone to catastrophic failures: child\#006, child\#008, adolescent\#002, and adult\#009

For the termination penalty experiment, we included child\#001, child\#003, adolescent\#002, adolescent\#008, adult\#008, and adult\#009. We used these patients as they contained a mix of regular and hard to control patients.

For seed progression experiments, we focused on a subset of patients we found reliably challenging for different models to control (in terms of risk and catastrophic failure rate). We included child\#001, child\#006, child\#008, adolescent\#002, adolescent\#003, adult\#001, and adult\#009 in this analysis.

\subsection{RL-Scratch on Adolescent\#002}\label{app:ao2}
RL-Scratch has a distinct form of degenerate behavior that occurs only in adolescent\#002, drastically lowering performance (see \textbf{Figure \ref{fig:full_risk_noma}}). We reliably observe that models trained on adolescent\#002 do not administer \textit{any} insulin, and thus achieve chronic hyperglycemia. This is because, unlike any other virtual patient, adolescent\#002 does not require any insulin to achieve blood glucose levels below 1000 mg/dL (the threshold at which we apply the hyperglycemic termination penalty) over a 10-day period. As a result of the large disparity between average rewards and the termination penalty, the network quickly learns to never administer insulin and the learned exploration rate collapses to 0. This approach can be easily avoided by warm-starting using an environment from another individual, as then the network has already learned to administer generally safe amounts of insulin. Adolescent\#002 spends, on average, 97\% of their time hyperglycemic under RL-Scratch, and only 39.2\% of time hyperglycemic under RL-Trans.

\end{document}